\documentclass{article}
\usepackage[utf8]{inputenc}
\usepackage{iclr2022_conference,times}

\usepackage{tikz}
\usepackage{tikz-3dplot}
\usepackage{bm}
\usepackage{wrapfig}
\usepackage{algorithm}
\usepackage{algpseudocode}
\usepackage{sidecap}
\sidecaptionvpos{figure}{t}
\usepackage{amssymb}
\usepackage{amsmath}
\usepackage[font=footnotesize]{subcaption}
\usepackage{dsfont}
\usepackage[colorlinks=true, citecolor=black, linkcolor=blue]{hyperref}
\usepackage[export]{adjustbox}
\usepackage{mathtools}
\usepackage{listings}
\usepackage{float}
\usepackage{booktabs}
\usepackage{floatflt}

\usepackage{eqparbox}
\usepackage{comment}

\usepackage{pgfplots}
\usetikzlibrary{calc}
\usetikzlibrary{shapes.geometric}
\usetikzlibrary{arrows}
\usetikzlibrary{fit}
\usetikzlibrary{positioning}
\usetikzlibrary{decorations.pathreplacing}
\usetikzlibrary{shadows}
\usetikzlibrary{backgrounds}

\definecolor{vred}{RGB}{225,86,44}
\definecolor{vblue}{RGB}{83,126,255}
\definecolor{vorange}{RGB}{0,203,133}

\DeclareMathOperator*{\argmax}{arg\,max}

\title{Graph Convolutional Memory using\\Topological Priors}

\author{
Steven D. Morad\thanks{Department of Computer Science and Technology, University of Cambridge, Cambridge, UK}, 
~Stephan Liwicki\thanks{Toshiba Europe Limited, Cambridge, UK}, 
~Ryan Kortvelesy\footnotemark[1], 
~Roberto Mecca\footnotemark[2],
~Amanda Prorok\footnotemark[1]\\
\texttt{\{sm2558,rk627,asp45\}@cam.ac.uk,}\\
\texttt{\{stephan.liwicki, roberto.mecca\}@crl.toshiba.co.uk}
}

\iclrfinalcopy
\begin{document}
\maketitle
\begin{abstract}
    Solving partially-observable Markov decision processes (POMDPs) is critical when applying reinforcement learning to real-world problems, where agents have an incomplete view of the world. We present graph convolutional memory (GCM)\footnote{GCM is available at \url{https://github.com/smorad/graph-conv-memory}}, the first hybrid memory model for solving POMDPs using reinforcement learning. GCM uses either human-defined or data-driven topological priors to form graph neighborhoods, combining them into a larger network topology using dynamic programming. We query the graph using graph convolution, coalescing relevant memories into a context-dependent belief. When used \textit{without} human priors, GCM performs similarly to state-of-the-art methods. When used \emph{with} human priors, GCM outperforms these methods on control, memorization, and navigation tasks while using significantly fewer parameters.
\end{abstract}
\section{Introduction}
Reinforcement learning (RL) was designed to solve \emph{fully observable} Markov decision processes (MDPs) \citep[Chapter~3]{sutton2018reinforcement}, where an agent knows its true state -- a property that rarely holds in the real world. Problems where agent state is ambiguous, incomplete, noisy, or unknown can be modeled as partially-observable MDPs (POMDPs). RL guarantees optimal policy convergence for POMDPs when a \emph{belief} is maintained over an episode \citep{Cassandra1994}. Storing information and retrieving it later for belief estimation is known as \emph{memory} \citep{moreno2018neural}.




Memory models in RL tend to be either general or task-specific. General memory comes from the field of sequence learning, with recurrent neural networks (RNNs), transformers, or memory augmented neural networks (MANNs) as notable examples. General memory assumes a sequential ordering, but makes no other assumptions about the inputs and can be applied to any POMDP. These models excel in supervised learning, but tend to be costly in terms of training time and number of parameters. RL exacerbates these problems with its sparse and noisy learning signal \citep{Beck}.

The substantial cost of training general memory drives many to design task-specific memory for RL applications, like \cite{Chaplot2020, Parisotto2017, Gupta2017a, Lenton2021} which build 2D or 3D maps for navigation, or \cite{li2018actor} which utilizes dosing information to inform a tree search over hospital patient states. Task-specific memory is built around human-defined prior knowledge, and aimed at solving a specific task. The downside of this memory is that it must be implemented by hand for each specific task. Many RL applications would benefit from task-specific memory, but the implementation is non-trivial: outside of the core RL research community there are chemists \citep{Zhou2017}, painters \citep{Huang2019}, or roboticists \citep{Morad2021} trying to solve field-specific problems using RL.

\begin{figure}
    \centering
    \scalebox{1.0}{\begin{tikzpicture}
    \tikzset{    
        box/.style = {%
            draw, 
            minimum width=3em, 
            minimum height=3em
        },
        nn/.style = {%
            draw, 
            circle,
            thick,
            black,
            inner sep=0pt,
            text width=1.5mm,
        },
        edge/.style = {
            -,
            black,
            semithick
        },
        node distance = 2em and 2em,
        mlp/.pic= {%
        \begin{scope}[every node/.append style={anchor=center}]
            \node[nn] (n1) {};
            \node[nn, above=0.3em of n1] (n0) {};
            \node[nn, below=0.3em of n1] (n2) {};
            
            \node[nn] (n3) at ([xshift=1em]$(n0)!0.5!(n1)$) {};
            \node[nn] (n4) at ([xshift=1em]$(n1)!0.5!(n2)$) {};
            
            \draw[edge] (n0.east) -- (n3.west);
            \draw[edge] (n1.east) -- (n3.west);
            \draw[edge] (n2.east) -- (n3.west);
            
            \draw[edge] (n0.east) -- (n4.west);
            \draw[edge] (n1.east) -- (n4.west);
            \draw[edge] (n2.east) -- (n4.west);
            
            \node[draw, fit=(n0) (n2) (n4), rounded corners=5pt, inner sep=2pt, thick] (box) {};
        \end{scope}
        },
        graph/.pic = {
        \begin{scope}[every node/.append style={anchor=center}]
            \node[gnode, red, anchor=center] (n0) at (0,0) {};
            \node[gnode, black] (n1) at (0.25, -0.5) {};
            \node[gnode, blue] (n2) at (0.25, 0.5) {};
            \node[gnode, orange, densely dashed] (n3) at (-0.5, 0.35) {};
            \draw[->] (n1) -- (n0);
            \draw[->] (n2) -- (n0);
        \end{scope}
        },
        gnode/.style = {
            draw,
            line width=1.5pt,
            circle
        },
    }
    
    \draw pic[local bounding box=g0](g0_) {graph};
    \node[draw, circle, inner sep=1pt, fit = (g0)] (g0_box) {};
    
    \draw pic[local bounding box=g1, right=8em of g0_box](g1_) {graph};
    \draw[<-] (g1_n3) -- (g1_n0);
    \draw[<-] (g1_n3) -- (g1_n2);
    \node[draw, circle, fit = (g1), inner sep=1pt] (g1_box) {};

    \node[draw, rounded corners=5pt, align=center] (es) at ($(g0)!0.5!(g1)$) {$N(o_t)$};
    
    \draw pic[local bounding box=gnn, right=of g1_box] (gnn) {mlp};
    \node[below=0.1em of gnn] (gnn_label) {GNN};
    
    \node[draw, text=black, thick, left=of g0, draw=orange] (b0) {$\cdot$};
    \node[draw, text=black, thick, below=0pt of b0, draw=orange] (b1) {$\cdot$};
    \node[above=0.1em of b0] {$o_t$};
    
    \node[draw, fit= (g0_box) (g1_box) (gnn) (gnn_label), label=below:GCM, rounded corners=5pt] (gcm) {};
    
    \node[draw, circle, right=of gnn] (bt) {$b_t$};
    
    \node[draw, rounded corners=5pt, right=of bt, minimum size=2em] (policy) {$\pi$};

    \draw[thick, ->, orange] (b0.south east) -- (g0_n3);
    \draw[thick, ->] (g0_box) -- (es);
    \draw[thick, ->] (es) -- (g1_box);
    \draw[thick, ->] (g1_box) -- (gnn);
    \draw[thick, ->] (gnn) -- (bt);
    \draw[thick, ->] (bt) -- (policy);

\end{tikzpicture}}
    \caption{GCM flowchart for an incoming observation $o_t$. GCM places $o_t$ as a node in a graph, and computes its neighborhood $N(o_t)$, and then updates the edge set. Task-specific topological priors are defined via the neighborhood. A convolutional GNN queries the graph for belief state $b_t$. A policy $\pi$ uses the belief for decision making. Compared to transformers or DNCs, GCM is architecturally simple.}
    \label{fig:gcm_flowchart}
\end{figure}
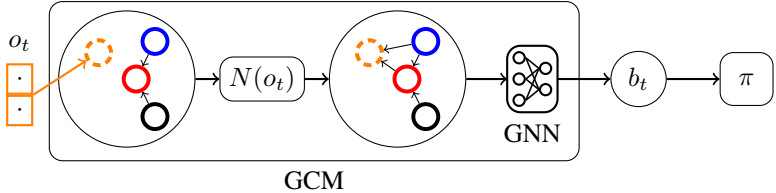

This paper is the first to propose a \emph{hybrid} memory model. Our memory model, termed Graph Convolutional Memory (GCM) is applicable to any partially-obserable RL task, but utilizes task-specific \emph{topological priors}. In other words, GCM builds a graph structure from topological priors, which are either human-designed or learned entirely from data. GCM has a simple interface, enabling users to develop topological priors suited to their specific tasks in a few lines of code (\autoref{sec:priors}). \autoref{fig:gcm_flowchart} presents an overview of our method. GCM builds a graph, and defines local neighborhoods using said priors, resulting in an expressive graph topology via dynamic programming. The edge structure induced by the graph provides interpretability, and facilitates simple debugging. We leverage the computational efficiency and representational power of graph neural networks (GNNs) to extract contextualized beliefs from the graph. GCM can be applied to any sequence learning problem, but in this paper we focus on RL. In our experiments, we show that GCM uses significantly fewer parameters than RNNs, MANNs, or transformers, but performs comparably without any human priors. With human-designed priors, GCM outperforms the general memory used in our experiments.
\section{Graph Convolutional Memory}
We model a POMDP following \cite{Kaelbling1998} with tuple $(\mathcal{S}, \mathcal{A}, \mathcal{T}, R, \Omega, \mathcal{O})$. At time $t$ we enter hidden state $s_t \in \mathcal{S}$ and receive observation $o_t \sim \mathcal{O}(s_t): S \rightarrow \Omega$. We sample action $a_t \in \mathcal{A}$ from policy $\pi$ and follow transition probabilities $\mathcal{T}(s_t, a_t): \mathcal{S} \times \mathcal{A} \rightarrow \mathcal{S}$ to the next state $s_{t+1}$, receiving reward $R(s_t, a_t): \mathcal{S} \times \mathcal{A} \rightarrow \mathbb{R}$. We learn $\pi$ to maximize the expected cumulative discounted reward subject to discount factor $\gamma$: $\mathbb{E} \left[ \sum_{t=0}^\infty \gamma^t R(s_t, a_t) \right]$. In an MDP, the policy uses the true state $\pi(s_t): \mathcal{S} \rightarrow \mathcal{A}$, but in a POMDP our policy uses the \emph{belief} state $\pi(b_t): \mathcal{B} \rightarrow \mathcal{A}$. Our goal in this paper is to construct a latent belief state $b_t$ using memory function $\mathrm{M}$ and memory state $m_t$:
\begin{equation}
    (b_t, m_t) = \mathrm{M}(o_t, m_{t-1}).
    \label{eq:mem}
\end{equation}

\subsection{Model Description}
\begin{figure}
    \centering
    \scalebox{0.8}{
               \begin{tikzpicture}[scale=1]
           \def\centerarc[#1](#2)(#3:#4:#5)
        { \draw[#1] ($(#2)+({#5*cos(#3)},{#5*sin(#3)})$) arc (#3:#4:#5); }
        \def\sft{2.1};
        \def\cthick{0.3em};
        \def\cshift{1.25em};
        \def\ctotal{2.8em}

            \node[circle, thick, minimum size=2.5em] (v0) at (0,2.25) {$o_t$};
            \node[draw, circle, vred, minimum size=2.5em, line width=\cthick] at (v0) {};
            
            \node[circle, thick,  minimum size=2.5em] (v1) at ($(v0) - (0, 1.25)$) {$o_{j}$};
            \node[draw, circle, vblue, minimum size=2.5em, line width=\cthick] at (v1) {};
            
            \node[circle, thick, minimum size=2.5em] (v2) at ($(v0) - (1, 1.25)$) {$o_i$};
            \node[draw, circle, vblue, minimum size=2.5em, line width=\cthick] at (v2) {};
            
            \node[circle, thick, minimum size=2.5em] (v11) at ($(v0) - (1, 2.5)$) {$o_{k}$};
            \node[draw, circle, vorange, minimum size=2.5em, line width=\cthick] at (v11) {};
            
            \coordinate (midpt0) at ($(v1)!0.5!(v2)$);
            \draw[dotted, ->, line width=1pt] (v1) -- (v0);
            \draw[dotted, ->, line width=1pt] -- (v2) -- (v0);
            \draw[dotted, ->, line width=1pt] (v11) -- (v2);
            
            \node[draw, rounded corners=.55cm,fit=(v1) (v2), inner sep=0.15cm, label] (agg) {};
            \node[left=0.1cm of v2] (n_label) {$N(o_t)$};
            \node[left=0.1cm of v11] (n1_label) {$N(o_i)$};
            \node[align=center, below=0.2cm of v1] (agg_label) {\texttt{agg}};

            \node[trapezium,draw,trapezium left angle=70, trapezium right angle=110, minimum height=2em] (w2) at ($(midpt0) + (\sft,0)$) {$W_2^1$};

            \node[trapezium,draw,trapezium left angle=70, trapezium right angle=110, minimum height=2em] (w1) at (w2 |- v0) {$W_1^1,b^1$};
            \coordinate (midpt1) at ($(w1)!0.5!(w2)$);
            
            \draw[->] (agg) -- (w2);
            \draw[->] (v0) -- (w1);
            
            \node[draw, circle] (sum) at ($(w1) + (1.5, 0)$) {$+$};
            \draw[->] (w1.east) -- (sum);
            \draw[->] (w2.east) to[out=0,in=-90] (sum);
            
            \node[draw, circle] (act) at ($(sum) + (1.0, 0)$) {$\sigma$};
            \draw[->] (sum) -- (act);
            
            \node[draw, dashed, label=below:Graph Layer 1,fit=(sum) (v0) (v1) (v11) (agg) (act) (n_label)] (fit0) {};

            \node[circle, minimum size=\ctotal] (v3) at ($(act) + (3.7, 0)$) {$z^1_t$};
            \draw[vred, line width=\cthick] ($(v3) + (0:\cshift)$) arc (0:180:\cshift);
            \draw[vblue, line width=\cthick] ($(v3) + (0:\cshift)$) arc (360:180:\cshift);
 
            \draw[->] (act) -- (v3);

            \node[circle, thick, minimum size=2.5em] (v4) at ($(v3) - (0, 1.25)$) {$z^1_{j}$};
            \node[draw, circle, vblue, minimum size=2.5em, line width=\cthick] at (v4) {};

            \node[circle, thick, minimum size=2.5em] (v5) at ($(v3) - (1, 1.25)$) {$z^1_{i}$};
            \draw[vblue, line width=\cthick] ($(v5) + (0:\cshift)$) arc (0:180:\cshift);
            \draw[vorange, line width=\cthick] ($(v5) + (0:\cshift)$) arc (360:180:\cshift);
            
            
            \node[circle, thick, minimum size=2.5em] (v55) at ($(v3) - (1, 2.5)$) {$z^1_{k}$};
            \node[draw, circle, vorange, minimum size=2.5em, line width=\cthick] at (v55) {};

            \coordinate (midpt1) at ($(v4)!0.5!(v5)$);
            \draw[dotted, ->, line width=1pt] (v4) -- (v3);
            \draw[dotted, ->, line width=1pt] (v5) -- (v3);
            \draw[dotted, ->, line width=1pt] (v55) -- (v5);
            
            \node[draw, rounded corners=.55cm,fit=(v4) (v5), inner sep=0.15cm] (agg1) {};
            \node[left=0.1cm of v5] (n_label1) {$N(o_t)$};
            \node[left=0.1cm of v55] (n1_label1) {$N(o_i)$};
            \node[align=center, below=0.2cm of v4] (agg_label1) {\texttt{agg}};
            
            \node[trapezium,draw,trapezium left angle=70, trapezium right angle=110, minimum height=2em] (w4) at ($(midpt1) + (\sft,0)$) {$W_2^2$};

            \node[trapezium,draw,trapezium left angle=70, trapezium right angle=110, minimum height=2em] (w3) at (w4 |- v3) {$W_1^2,b^2$};
            \coordinate (midpt1) at ($(w3)!0.5!(w4)$);
            
            \draw[->] (agg1.east |- w4) to[out=0,in=180] (w4);
            \draw[->] (v3) -- (w3);
            
            \node[draw, circle] (sum1) at ($(w3) + (1.5, 0)$) {$+$};
            \draw[->] (w3.east) -- (sum1);
            \draw[->] (w4.east) to[out=0,in=-90] (sum1);
            
            \node[draw, circle] (act1) at ($(sum1) + (1.0, 0)$) {$\sigma$};
            \draw[->] (sum1) -- (act1);
            
            \node[draw, dashed, label=below:Graph Layer 2,fit=(sum1) (v3) (v4) (v5) (v55) (agg1) (act1) (n_label1)] (fit1) {};
            
            \node[circle, minimum size=\ctotal] (v6) at ($(act1) + (1.25, 0)$) {$z^2_t$};
            \draw[vorange, line width=\cthick] ($(v6) + (0:\cshift)$) arc  (0:360:\cshift);  
            \draw[vblue, line width=\cthick] ($(v6) + (0:\cshift)$) arc (0:240:\cshift);
            \draw[vred, line width=\cthick] ($(v6) + (0:\cshift)$) arc (0:120:\cshift);
            \node[below=0.1cm of v6] (s_label) {$b_t$};

            \draw[->] (act1) -- (v6);
        \end{tikzpicture}
     }
    \caption{The two-layer 1-GNN used in all our experiments. Colors denote mixing of vertex information and dashed lines denote directed edges, forming neighborhoods $N(o_t), N(o_i)$. The current observation $o_t$ and aggregated neighboring observations $o_{i}, o_{j}$ pass through fully-connected layers $(W_1^1, b^1), (W_2^1)$ before summation and nonlinearity $\sigma$, resulting in the first hidden state $z^1_t$ (\autoref{eq:gc}). We repeat this process at $o_{j},o_{k}, o_{l}$ to form hidden states $z^1_j, z^1_k, z^1_l$. The second layer combines embeddings of the first layer and the second-layer hidden state $z_t^2$ is output as the belief state $b_t$. Additional layers increase the GNN receptive field.}
    \label{fig:graphconv_viz}
\end{figure}
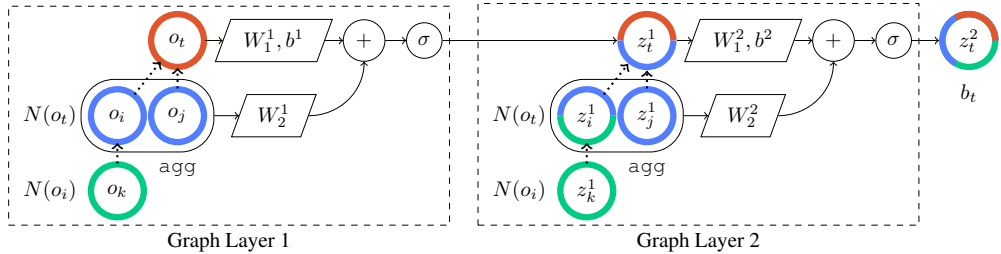

We implement GCM following \autoref{eq:mem} using \hyperref[alg:GCM]{Alg.~\ref*{alg:GCM}}. GCM stores a collection of experiences over an episode, with each experience represented by an observation vertex $o$ and associated neighborhood $N(o)$. We query the set of experiences using a graph neural network (GNN) to produce a context-dependent belief $b_t | o_t$.
\begin{floatingfigure}[r]{0.53\linewidth}
    \vspace{-1em}
    \begin{minipage}{0.53\linewidth}
        \begin{algorithm}[H]
            \caption{Graph Convolutional Memory}
            \label{alg:GCM}
            \begin{algorithmic}[1]
                \small
                \Procedure{M}{$o_t, m_{t-1}$}
                    \State $V, E \gets m_{t-1}$ \Comment{Unpack memory}
                    \State $V \gets V \cup o_t$ \Comment{Add observation}
                    \State $E \gets E \cup \{(o_t, o_i)\}_{i \in N(o_t)}$ \Comment{Update edges}
                    \State $Z \gets \mathrm{GNN}_{\theta} (V, E)$
                    \Comment{Get embedding}
                    \State $b_t \gets Z\left[ t \right]$ \Comment{At current vertex}
                    \State $m_t \gets V, E$ \Comment{Pack into memory}
                    \State \textbf{return} $b_{t}, m_t$ \Comment{Belief and memory}
                \EndProcedure
            \end{algorithmic}
        \end{algorithm}
    \end{minipage}
    \vspace{1em}
\end{floatingfigure}
In detail, at time $t$, we insert vertex $o_t$ into the graph, producing $m_t = (V_t,E_t)$ where $V_t = (o_1, \dots o_t)$ and $E_t: V_t \times V_t$. We determine the neighborhood $N(o_t)$ using \emph{topological priors} defined in \autoref{sec:priors}, and update the edges following:
\begin{align}
    E_t = E_{t-1} \cup \{(o_t, o_i) \mid i \in N(o_t)\}
\end{align}
We query the graph for context-dependent information using a GNN with layers $h \in \{1 \dots \ell\}$. We convolve over $o_1, \dots o_t$ to produce hidden representations $z_1^h, \dots z_t^h$ for each hidden layer, propagating information from the $h$\textsuperscript{th}-degree neighbors of $o_t$ into $z_t^h$. After collecting and integrating data across the $\ell$\textsuperscript{th}-degree neighborhood, we output $z_t^\ell$ as the belief. This provides a mechanism for fast and relevant feature aggregation over  memory graphs, depicted in \autoref{fig:graphconv_viz}. 

As an example, let us consider a visual navigation task where the neighborhood consists of the previous observation index $N(o_t) = \{t-1\}$. Let $o_1, o_2, o_3$ represent ``chair'', ``wall'', and ``table'', respectively. The first GNN layer combines $o_1, o_2$ into a ``chair-room'' embedding and $o_2, o_3$ into a ``table-room'' embedding. The second layer combines ``chair-room'' and ``table-room'' into ``dining-room'', and outputs ``dining-room'' as the belief.
We found the 1-GNN defined in \cite{Morris2019} empirically outperformed graph isomorphism networks \citep{Xu2019} and the original graph convolutional network \citep{Kipf2017}. GCM can utilize any GNN, but our GNNs are built from the 1-GNN convolutional layer defined as:
\begin{align}
    z_t^{h} &= \sigma \left[ W_1^{h} z_t^{h-1} + b^{h} + W_2^{h} \texttt{agg} \left( \left\{ z^{h-1}_i | i \in N(o_t) \right\} \right) \right]
    \label{eq:gc}
\end{align}
with $\sigma$ representing a nonlinearity and $z_t^0 = o_t, z_i^0 = o_i $ for the base case. At each layer $h$, weights and biases $W_1^h, b^h$ produce a root vertex embedding while $W_2^h$ generates a neighborhood embedding using vertex aggregation function \texttt{agg}. Separate weights allow the 1-GNN to weigh each $h$\textsuperscript{th}-degree neighborhood's contribution to the belief, ignoring the neighborhood and decomposing into an MLP for empty or uninformative neighborhoods. The root and neighborhood embeddings are combined to produce the layer embedding $z_t^h$ (\autoref{fig:graphconv_viz}). Notice, the weights $W_1^h, b^h$ in \autoref{eq:gc} form an MLP, so GCM does not require an MLP preprocessor like other memory models \citep{Mnih2016}.

\subsection{Topological Priors}
\label{sec:priors}
We use the shorthand $N(o_t)$ to define the open neighborhood of $o_t$ over vertices $V_t$, in edge-list format. We compute $N(o_t)$ using the union of one or more \emph{topological priors} $\Phi_i: \Omega^{t} \rightarrow 2^{V_{t-1}}$, as in
\begin{align}
    \label{eq:neighbors}
    N(o_t)&: V \rightarrow 2^{V_{t-1}} \coloneqq \bigcup_{i=1}^{k} \Phi_i(V_{t}). 
\end{align}
Breaking down the graph connectivity problem into easier neighborhood-forming subtasks is a form of \emph{dynamic programming}. The priors are task-specific, because for different tasks, we often want different connectivity. For example, in control problems, associating memories temporally could be beneficial in learning a dynamics model. In navigation, spatial connectivity could aid with loop closures. We have implemented spatial, temporal, latent similarity, and other topological priors in \autoref{tab:priors}, but GCM can utilize any mapping from vertices to a neighborhood. We provide task-specific examples for medicine and aerospace in \autoref{sec:eg_priors}.

\begin{table}
\centering
\footnotesize
\renewcommand{\arraystretch}{0.1}
\begin{tabular}{@{}m{0.7\linewidth}m{0.28\linewidth}}
     Prior Description & $\Phi(V)$ Definition  \\
     \hline
    \textbf{Empty:} $o_t$ has no neighbors and GCM decomposes into an MLP. &  \begin{equation} 
        \emptyset 
        \label{eq:p_none}
    \end{equation}\\
    
    \textbf{Dense:} Connects $o_t$ to all other observations $o_{1} \dots o_{t-1}$. &
    \begin{equation}
        \{1, 2, \dots t-1\}
        \label{eq:p_dense}
    \end{equation}\\
    
    \textbf{Temporal:} Similar to the temporal prior of an LSTM, where each observation $o_t$ is linked to some previous $t-c$ observation. &
    \begin{equation}
        \{t-c\}
        \label{eq:p_temporal}
    \end{equation}\\
    
    \textbf{Spatial:} Connects observations taken within $c$ meters of each other, useful for problems like navigation. Let $\mathrm{p}(\cdot)$ extract the position from an observation. &
    \begin{equation}
        \left\{i \, \middle\vert \, \begin{array}{c}
        || \mathrm{p}(o_i) - \mathrm{p}(o_t) ||_2 \leq c \\
        \textrm{ and } 0 < i < t \end{array}\right\}
        \label{eq:p_distance}
    \end{equation}
    \\
    
    \textbf{Latent Similarity:} Links observations in a non-human readable latent space (e.g.~autoencoders). Various measures like cosine or $L_2$ distance may be used depending on the space. $\mathrm{e}$ is an encoder function, $\mathrm{d}$ is a distance measure, and $c$ is user-defined. &
    \begin{align}
        \left\{i \, \middle\vert \, \begin{array}{c} \mathrm{d}\left( \mathrm{e} \left(o_i \right), \mathrm{e} \left( o_t \right) \right) < c\\
        \textrm{ and } 0 < i < t \end{array} \right\}
        \label{eq:p_vae}
    \end{align}\\
    \\
    
    \textbf{Identity:} Connects observations where two values are identical, useful in discrete domains where inputs are related. $\mathrm{a}$, $\mathrm{b}$ are indexing functions ($\mathrm{a} = \mathrm{b}$ may hold).  &
    \begin{equation}
         \left\{ i \, \middle\vert \, \begin{array}{c}\mathrm{a}(o_i) - \mathrm{b}(o_t) = 0\\ 
         \textrm{ and } 0 < i < t \end{array} \right \}
        \label{eq:p_identity}
    \end{equation}\\
    
    \hline
\end{tabular}
\caption{Knowledge-based priors for GCM}\label{tab:priors}
\end{table}

\paragraph{Learning Topological Priors}
We stress the importance of human knowledge in forming the graph topology, but there are cases where we have no information about the problem, or where designing a topological prior is non-trivial. A na\"ive formulation of prior learning via gradient descent is challenging due to a large number of possible edges represented by boolean values. One option is to use a fully-connected graph with weighted edges similar to \cite{Velickovic2018}, but this would hinder interpretability of the graph \citep{Jain2019} and be computationally expensive. 

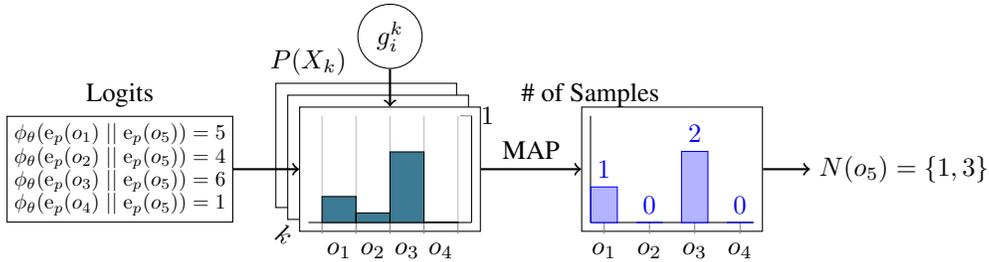
\begin{figure}
    \centering
    \pgfmathdeclarefunction{gauss}{2}{%
  \pgfmathparse{1/(#2*sqrt(2*pi))*exp(-((x-#1)^2)/(2*#2^2)) + 1/(#2*sqrt(2*pi))*exp(-((x-#1+2)^2)/(8*#2^2))}%
}

\pgfplotstableread{
X Y
0 0.2436
1 0.0896
2 0.6623
3 0.0045
}\datatable

\begin{tikzpicture}
\tikzset{
   cascaded/.style = {%
        general shadow = {%
            shadow scale = 1,
            shadow xshift = -2ex,
            shadow yshift = 2ex,
            draw=black,
            fill = white,
        },
        general shadow = {%
            shadow scale = 1,
            shadow xshift = -1ex,
            shadow yshift = 1ex,
            draw=black,
            fill = white
        },
        draw,
        fill = white
    },
}
\begin{axis}[
    ybar interval,
    domain=1:5, 
    samples=100,
    ymax=1,ymin=0,
    axis x line*=bottom,
    axis y line*=right,
    every axis y label/.style={at=(current axis.above origin),anchor=south},
    height=3cm, width=3.75cm,
    xticklabels={$o_1$,$o_2$,$o_3$,$o_4$},
    xtick={1,2,3,4,5},
    ytick={1},
    ylabel={$P(X_k)$},
    y label style={at={(axis description cs:0,1.3)}},
    name=p1,
    title style={yshift=1.6ex}
]
  \addplot[fill=cyan!50!black] coordinates{ (1, 0.2436) (2, 0.0896) (3, 0.6623) (4, 0.0045) (5, 0) };
\end{axis}

\node[rotate=-30] at (-.32cm, -.15cm) {$k$};

\begin{scope}[on background layer]
    \node[draw, fit=(p1), cascaded, fill=white] (pp1) {};
\end{scope}

\begin{axis}[
    ybar, 
    ymax=3,ymin=0, 
    at={(3.75cm,0)},
    no markers, domain=1:5, samples=100,
    axis lines*=left, 
    ylabel={\# of Samples},
    every axis y label/.style={at=(current axis.above origin),anchor=south},
    height=3cm, width=3.75cm,
    xticklabels={$o_1$,$o_2$,$o_3$,$o_4$}, ytick=\empty,
    xtick={1,2,3,4},
    nodes near coords,
    y label style={at={(axis description cs:-0,1)}},
    name=p2,
    title style={yshift=1.6ex}
]
\addplot coordinates { (1, 1) (2, 0) (3, 2) (4, 0)  };
\end{axis}

\begin{scope}[on background layer]
    \node[draw, fit=(p2), fill=white] (pp2) {};
\end{scope}

\node[draw, left=1 of p1, label=above:Logits, align=center, scale=0.8] (logits) {
$\phi_\theta(\mathrm{e}_p(o_1) \mid \mid \mathrm{e}_p(o_5))=5$\\ 
$\phi_\theta(\mathrm{e}_p(o_2) \mid \mid \mathrm{e}_p(o_5))=4$\\ 
$\phi_\theta(\mathrm{e}_p(o_3) \mid \mid \mathrm{e}_p(o_5))=6$\\ 
$\phi_\theta(\mathrm{e}_p(o_4) \mid \mid \mathrm{e}_p(o_5))=1$}; 

\node[draw, circle, above=0.5 of pp1] (noise) {$g^k_i$};

\node[right=0.75 of p2, align=center] (neighbors) {$N(o_5) = \{1,3\}$};

\draw[->, thick] (logits) -- (pp1);
\draw[->, thick] (noise) -- (pp1);
\draw[->, thick] (pp1) -- (pp2) node[above, midway] {MAP};
\draw[->, thick] (pp2) -- (neighbors);
\end{tikzpicture}
    \caption{Generating one to $K$ edges in a non-deterministic manner, using our learned topological prior. We present an example of the learned prior with $K=3$ at $t=5$. An MLP computes logits over previous vertices to produce three distributions $X_k; k \in \{1,2,3\}$, perturbed with Gumbel noise $g^k_i$. We compute the MAP for each $X_k$, resulting in three samples which form a two-edge neighborhood $N(o_5)$ containing vertices $o_1, o_3$. This process is fully-differentiable and trained end-to-end with GCM.}
    \label{fig:learned_viz}
\end{figure}

Instead, we propose to learn a probability distribution over all edges, from which we sample to explore the large edge space. We do not know the ideal neighborhood size, so we must learn this as well. The Gumbel-Softmax Estimator \citep{Jang2016} enables differentiable sampling from a categorical distribution using the reparameterization trick from \cite{Kingma2014}. We leverage this to build a multinomial distribution across all possible edges, and use the maximum a posteriori (MAP) estimate to form the neighborhood (\autoref{fig:learned_viz}).





Assume we want to sample between one and $K$ edges at time $t$ for the neighborhood $N(o_t)$. First, we compute logits $l_i$ for all possible edges $(o_i, o_t), i \in \{1, \dots t-1\}$ in \autoref{eq:logits}, using a neural network $\phi_\theta$, positional encoding $\mathrm{e}_p$ from \cite{Vaswani2017}, and concatenation operator $\mid \mid$. Then, we sample Gumbel noise $g^k_i, k \in \{1,\dots K\}$ using the inverse CDF of the Gumbel distribution and uniform random sampling $\mathcal{U}(0,1)$ in \autoref{eq:noise}. Using Gumbel noise $g^k_i$, we build a multinomial distribution from $K$ Gumbel-Softmax distributions $X_k$ (\autoref{eq:multi}) which we ``sample'' by taking the MAP ($\textrm{argmax}$) (\autoref{eq:lprior}). 
Note that we are not actually sampling from a single categorical distribution, but taking the MAP over many perturbed categorical (i.e.~multinomial) distributions. This approach is formalized as follows: 
\begin{align}
    \label{eq:logits}
    l_i &= \phi_\theta \left( \mathrm{e}_p(o_i) \mid \mid \mathrm{e}_p(o_t) \right)\\
    \label{eq:noise}
    g^k_i &= -\log(-\log u); u \sim \mathcal{U}(0,1)\\
    \label{eq:multi}
    P(X_k)
    &= \left\{
    \frac{
        \exp{(\log l_i + g^k_i)}
    }{
        \sum_{j=1}^{t-1} \exp{(\log l_j + g^k_i)}
    }
    \, \middle\vert \,  1 \leq i < t 
    \right\}\\
    \label{eq:lprior}
    \Phi(V_t) & \coloneqq \left\{\argmax{P(X_k)} \mid 1 \leq k \leq K \right\}.
\end{align}
Random sampling helps our method explore more neighborhood possibilities while still providing boolean values for edges. Since we are sampling with replacement, the neighborhood size can vary from 1 to $K$, depending on the kurtosis of the learned distribution. For tasks where information is replicated over many observations, \autoref{eq:lprior} can learn a flat distribution, sampling many distinct edges and building a large neighborhood. In cases where a few key observations contain salient information, we can learn a peaked distribution to more reliably sample these edges.

We do not run into the interpretability trap of transformers explained in \cite{Jain2019}, because we learn a distribution of edges rather than attention weights over the edges. In a transformer, an infinitesimal perturbation in attention weights can cause a significant change in model output. Our model learns a distribution, where an infinitesimal perturbation corresponds to an infinitesimal shift of probability mass. An infinitesimally-shifted distribution will not significantly impact the drawn samples, or in turn the learned prior output. Furthermore, the learned prior should be robust to noisy distributions, since the distributions themselves are constantly perturbed with Gumbel noise during learning. We note this sampling methodology might also be useful for designing differentiable hard and sparse self-attention in transformers.

\section{Experiments}
\begin{figure}
    \centering
    \begin{subfigure}{0.19\linewidth}
        \centering
        \includegraphics[height=0.9\linewidth]{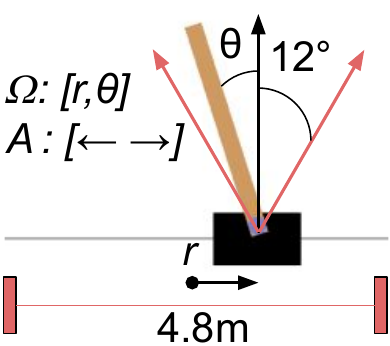}
        \caption{Partially obs. cartpole}
        \label{fig:sub_cartpole}
    \end{subfigure}
    \hfill
    \begin{subfigure}{0.39\linewidth}
        \centering
        \scalebox{0.65}{\begin{tikzpicture}
    \node[draw, minimum height=2em, minimum width=1.25em] (0) {2};
    \node[above=.15cm of 0] (_0) {0};
    \node[draw, minimum height=2em, minimum width=1.25em, right=.5cm of 0] (1) {};
    \node[above=.15cm of 1] (_1) {1};
    \node[draw, minimum height=2em, minimum width=1.25em, right=.5cm of 1] (2) {2};
    \node[above=.15cm of 2] (_2) {2};
    \node[draw, minimum height=2em, minimum width=1.25em, right=.5cm of 2] (3) {};
    \node[above=.15cm of 3] (_3) {3};
    \node[draw, minimum height=2em, minimum width=1.25em, right=.5cm of 3] (4) {};
    \node[above=.15cm of 4] (_4) {4};
    \node[draw, minimum height=2em, minimum width=1.25em, right=.5cm of 4] (5) {1};
    \node[above=.15cm of 5] (_5) {5};
    
    \node[below=0.5cm of 3, align=center] (ptr) {Pointer};
    \node[below=0.1cm of ptr, align=center] (act) {$\leftarrow$ flip $\rightarrow$};
    \draw[->, line width=1pt] (ptr) -- (3);
    \node[draw, dashed, label=below:$A$,fit=(act)] (act_agg) {};
    
    \node[below=0.5cm of 5, align=center] (obs) {$p: \left[\emptyset, 3\right]$\\$f:\left[1,5\right]$\\$a_{t-1}:\leftarrow$};
    \node[draw, dashed, label=below:$o_t$,fit=(obs)] (obs_agg) {};

    \draw[->, dashed] (3) -- (obs_agg);
    \draw[->, dashed] (5) -- (obs_agg);
\end{tikzpicture}}
        \caption{Concentration card game}
        \label{fig:sub_memory}
    \end{subfigure}
    \hfill
    \begin{subfigure}{0.39\linewidth}
        \centering
        \includegraphics[width=0.45\linewidth,angle=90]{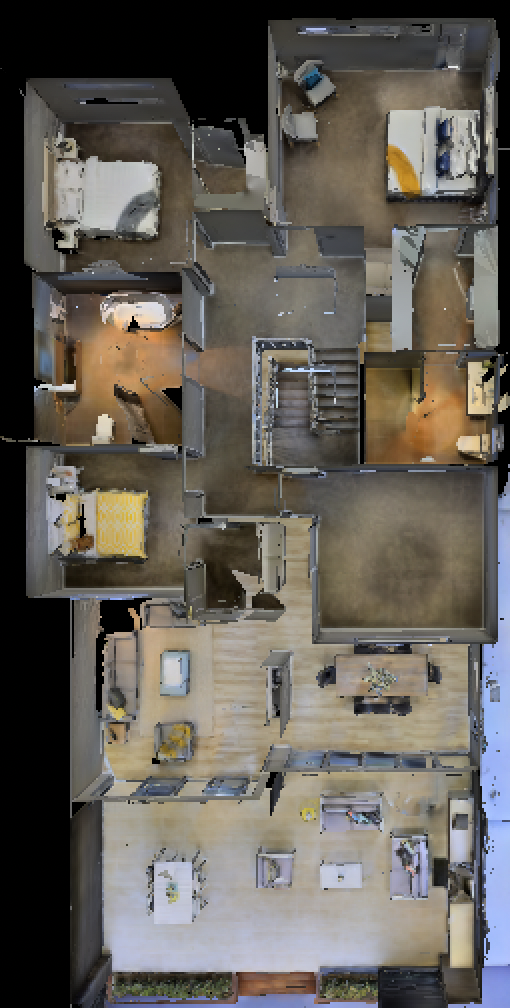}
        \caption{Habitat 3D simulation}
        \label{fig:sub_nav}
    \end{subfigure}
    \caption{Visualizations of our experiments. (\subref{fig:sub_cartpole}) The classic cartpole control problem, but where $\dot{r}, \dot{\theta}$ are hidden. (\subref{fig:sub_memory}) An example state from the long-term non-sequential recall environment with six cards. The observation space $o_t$ contains the value and index of pointer card $p$ and last flipped card $f$, as well as previous action $a_{t-1}$. (\subref{fig:sub_nav}) The top-down view of the 3D scene used in our navigation experiment.}
    \label{fig:exp_overview}
\end{figure}

\begin{wrapfigure}{r}{.4\textwidth}
    \begin{minipage}{\linewidth}
        \includegraphics[width=\linewidth]{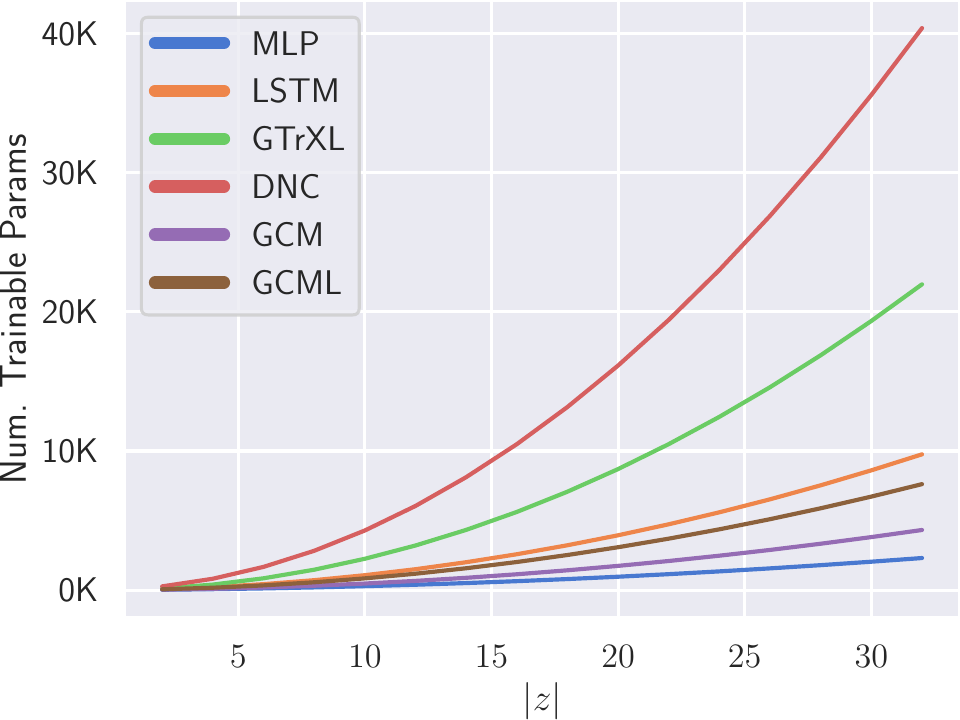}
        \subcaption{}
        \label{fig:sub_params}

    \small
    \begin{tabular}{p{0.12\linewidth} p{0.78\linewidth}}
            Model &  Meaning of $|z|$ \\
            \hline
            MLP & Layer size\\
            LSTM & Size of hidden and cell states\\
            GTrXL & Size of the attention head and position-wise MLP\\
            DNC & LSTM size, word width, and number of memory cells\\
            GCM & Size of the graph layers\\
            GCML & Size of graph layers and $\phi_\theta$ layers\\\hline
        \end{tabular}
        \subcaption{}
        \label{tab:params}
    \end{minipage}
    \caption{(\subref{fig:sub_params}) The number of trainable parameters per memory model, based on the hidden size $|z|$. GCM uses much fewer parameters than other memory models. (\subref{tab:params}) The meaning of $|z|$ with respect to each memory model, as used in all our experiments.}
    \label{fig:params}
\end{wrapfigure}

We evaluate GCM on control (\autoref{fig:sub_cartpole}), non-sequential long-term recall (\autoref{fig:sub_memory}), and indoor navigation (\autoref{fig:sub_nav}). We run three trials for each memory model across all experiments and report the mean reward per train batch with a 90\% confidence interval. We test six contrasting models, and base our evaluation on the hidden size of the memory models, denoted as $|z|$ in \autoref{fig:params}. Nearly all hyperparameters are Ray RLlib defaults, tuned for RLlib's built-in models.\footnote{The MLP, LSTM, DNC, and GTrXL are standard Ray RLlib \citep{Liang2018} implementations written in Pytorch \citep{Paszke2019}. We implement GCM using Pytorch Geometric \citep{Fey2019}, and integrate it into RLlib.} \autoref{sec:exp_details} defines all training environment details and \autoref{sec:hyperparams} contains a complete list of hyperparameters.

We compare GCM to an MLP and four alternative memory models in all our experiments. The \textbf{MLP} model is a two-layer feed-forward neural network using $\tanh$ activation. It has no memory, and forms a performance lower bound for the memory models. The \textbf{LSTM} memory model is a MLP followed by an LSTM cell, the standard model for solving POMDPs \citep{Mnih2016}. \textbf{GTrXL} is a MLP followed by a single-head GRU-gated transformer XL. The \textbf{DNC} is an MLP followed by a neural computer with an LSTM-based memory controller. Our memory model, \textbf{GCM}, uses a two-layer 1-GNN using $\tanh$ activation with sum (cartpole and concentration) and mean (navigation) neighborhood aggregation. The \textbf{GCML} is the GCM with the learned topological prior (\autoref{eq:lprior}) with $K=5$, and $\phi_\theta$ as a two-layer MLP using ReLU and LayerNorm \citep{Ba2016}.

\begin{figure}
    \centering
    \includegraphics[width=0.95\linewidth]{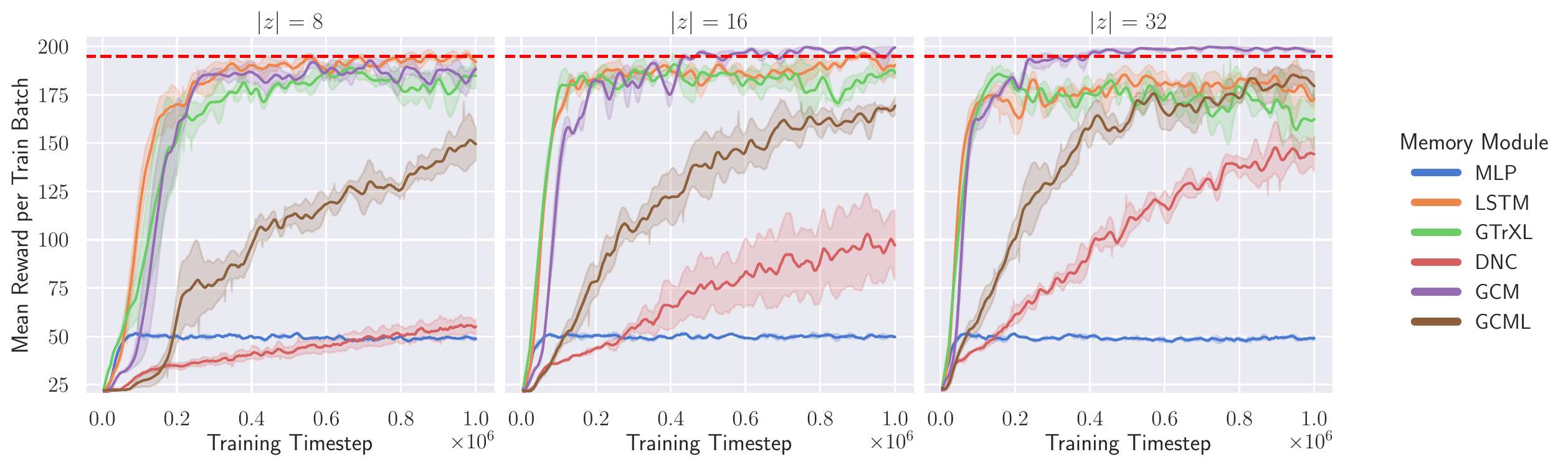}
    \caption{Stateless cartpole, where the agent must derive velocity from past observations. OpenAI considers fully-observable cartpole solved at a reward of 195 (dashed red line), which only GCM can reliably reach in partially-observable cartpole. Results represent the mean and 90\% confidence interval over three trials.}
    \label{fig:cartpole}
\end{figure}

\begin{figure}
    \centering
    \includegraphics[width=0.95\linewidth]{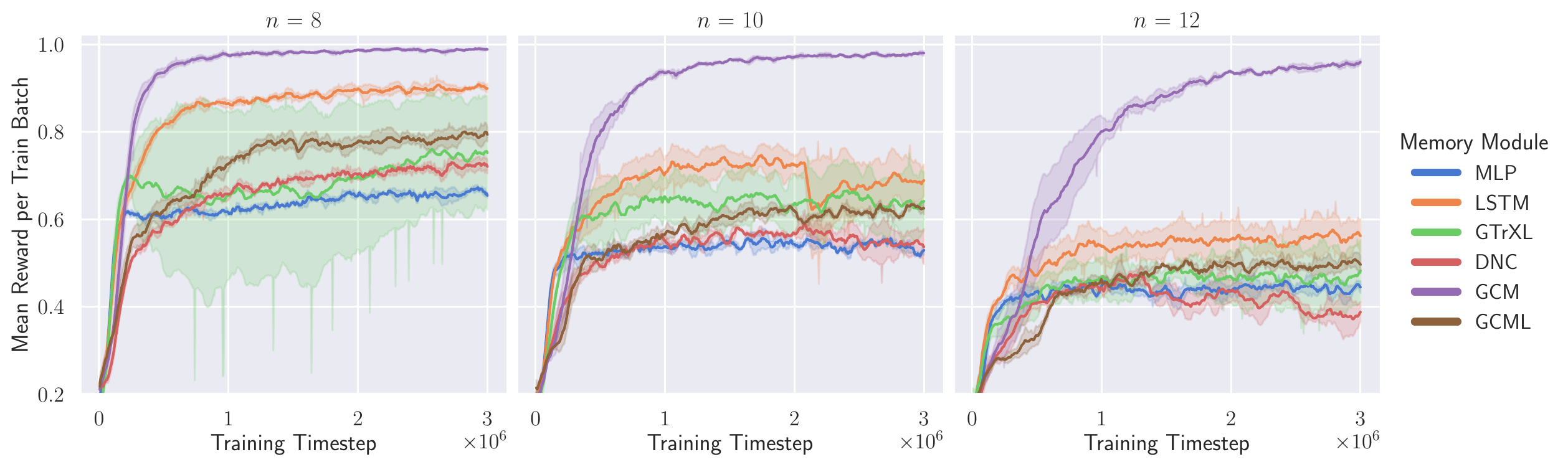}
    \caption{Results from concentration with hidden size $|z|=32$, where $n$ is the number of cards. This tests the agent's long-term non-sequential memory. The agent receives a small reward for matching a pair of cards, receiving a cumulative reward of one for matching all pairs. Episode lengths are 50, 75, and 100 respectively. Results are averaged over three trials and the shaded area represents the 90\% confidence interval.}
    \label{fig:memory}
\end{figure}

\begin{figure}
    \centering
    \includegraphics[width=0.95\linewidth]{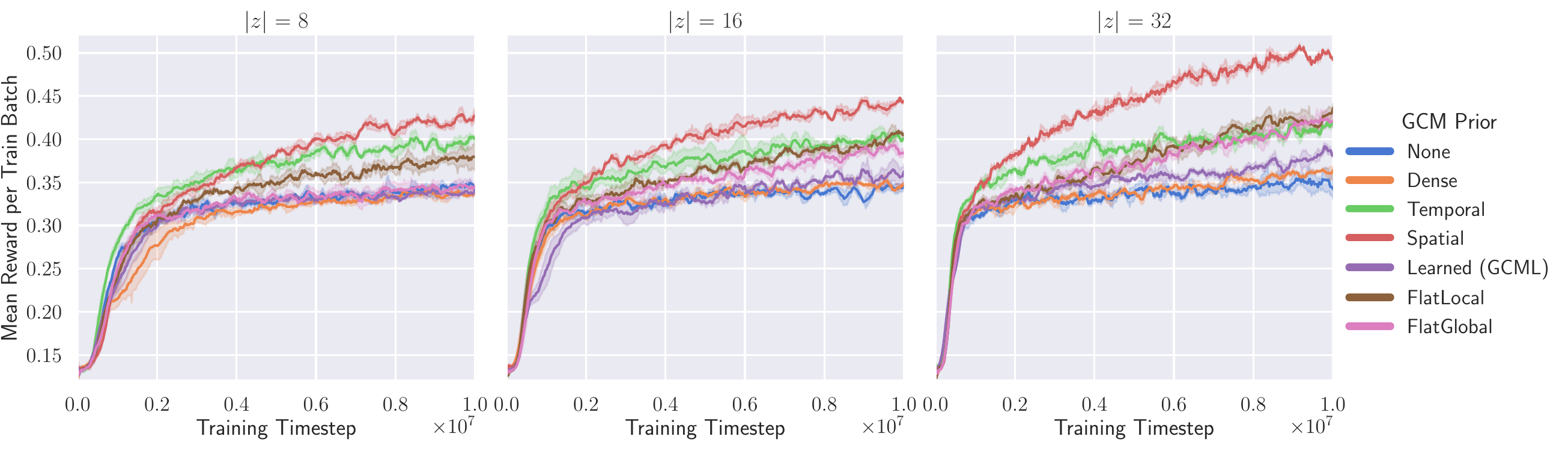}
    \caption{We compare various GCM priors across hidden sizes $|z|$ for the navigation problem. Since navigation is a spatial problem, the spatial prior performs best. This shows the importance of selecting good priors. Results are averaged over three trials and the shaded area represents the 90\% confidence interval.}
    \label{fig:nav_edge}
\end{figure}

\begin{figure}
    \centering
    \includegraphics[width=0.95\linewidth]{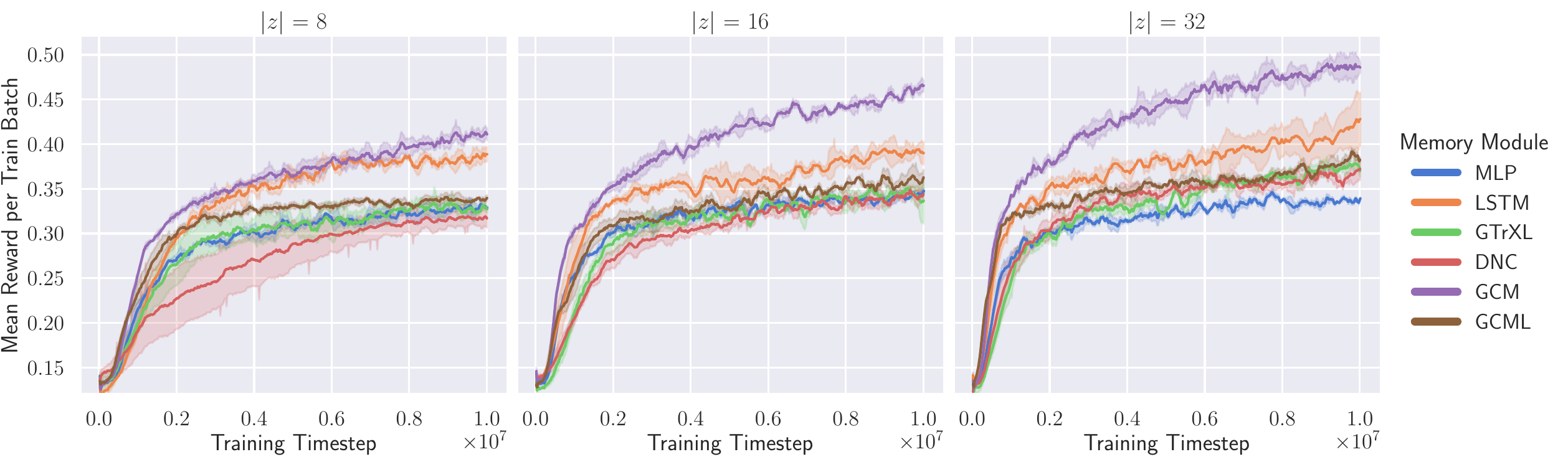}
    \caption{We compare GCM to other memory baselines for the navigation problem. $|z|$ denotes the hidden size used across all models. Results are averaged over three trials and the shaded area represents the 90\% confidence interval.}
    \label{fig:nav}
\end{figure}

\paragraph{Partially Observable Cartpole}
Our first experiment evaluates memory in the control domain. We use a partially observable form of cartpole-v0, where the observation corresponds to positions rather than velocities (\autoref{fig:sub_cartpole}). We optimize our policy using proximal policy optimization (PPO) \cite{schulman2017proximal}. The equations of motion for the cartpole system are a set of second-order differential equations containing the position, velocity, and acceleration of the system \citep{Barto1983}. Using this information, we use GCM with temporal priors, i.e., $N(o_t) = \{t-1, t-2\}$ (\autoref{tab:priors}) and present results in \autoref{fig:cartpole}.
\paragraph{Concentration Card Game}
Our next experiment evaluates non-sequential and long-term recall with the concentration card game.\footnote{Rules for concentration are available at: \url{https://en.wikipedia.org/wiki/Concentration_(card_game)}} Unlike reactionary cartpole, this experiment tests memorization and recall over longer time periods. We vary the number of cards $n \in \{8, 10, 12\}$ with episodes lengths of $50, 75, 100$ respectively. All models have $|z|=32$ and train using PPO. We use GCM with temporal priors for short-term memory and an additional value identity prior between the face-up card and the card at the pointer, using function $v : \Omega \rightarrow \mathbb{N}$:
\begin{equation}
    N(o_t) = \{t-1,t-2\} \cup \{i \vert v(o_t) = v(o_i)\}.
\end{equation}
In other words, when GCM flips a card face up, it recalls if it has seen that card in the past. We present the results in \autoref{fig:memory}.
\paragraph{Navigation}
The final experiment evaluates spatial reasoning with a navigation task. We use the Habitat simulator with the validation scene from the 2020 Habitat Challenge (\autoref{fig:sub_nav}). We train for 10M timesteps using IMPALA \citep{Espeholt2018}, examining $z \in \{8, 16, 32\}$ across all models. \autoref{fig:nav_edge} is an ablation study across multiple topological priors and GCM setups. We evaluate the effectiveness of empty, dense, temporal, spatial, and learned priors (formally defined in \autoref{tab:priors}). We also examine whether the larger receptive field induced by the second-degree neighborhood is helpful with the FlatLocal and FlatGlobal entries. The FlatLocal GCM uses the spatial prior and replaces the second GNN layer with a fully-connected layer, restricting GCM to its first-degree neighborhood. The FlatGlobal GCM is the FlatLocal GCM, but with an increased-distance spatial prior such that the first-degree neighborhood includes the first and second-degree neighborhoods of the spatial entry. \autoref{fig:nav} compares GCM with learned and spatial topological priors against other memory baselines. 
\section{Discussion}
The versatility of GCM compared to other models stems from its representation of experiences as a graph. This allows it to access specific observations from the past, bypassing the limited temporal range of LSTM. By using a multilayer GNN to reason over this graph of experiences, GCM can build embeddings hierarchically, unlike transformers. The importance of hierarchical reasoning is demonstrated experimentally in \autoref{fig:nav_edge}, where the GCM outperforms the FlatGlobal GCM, which merges the first and second-degree neighborhood from the spatial prior into a first-degree neighborhood. To build some intuition about why this is the case, consider an observation graph in a navigation experiment. Reasoning over this structure hierarchically can break down the task into manageable subtasks. The first GNN layer can fuse neighborhood viewpoints to represent local surroundings, and the second layer can combine its neighborhood of local surroundings into regions for planning. On the other hand, in a flat representation, information about the relationships between individual observations is lost. The flat model receives unstructured observations, and must learn to differentiate nearby observations from distant ones.

Like \cite{Beck}, we find sequence learning is much harder in RL than supervised learning. This is particularly clear in \autoref{fig:memory}, where introducing one more pair of cards decreases reward. Although general memory models can learn optimal policies in theory, this was not the case given our timescales. The LSTM performs well but does not reliably solve (i.e. reach 195 reward) stateless cartpole, even with small 2-dimensional observation and action spaces, and a large number of inner and outer PPO iterations (\cite{Heess2015}, \autoref{fig:cartpole}). Even though transformers significantly outperform LSTMs in supervised learning \citep{Vaswani2017}, their added complexity seems to hinder them in RL, at least at single-GPU scales. The memory search space over all past observations is huge, and determining which observations are useful greatly reduces what the memory model must learn. 

GCM's graph structure can utilize external information about which experiences are relevant, greatly reducing the search space. Human intuition is an incredibly useful tool that cannot be easily leveraged by transformers, RNNs, or MANNs. This is the key contribution of our work -- a prior defined by a few lines of code can significantly boost RL performance. GCM provides an easy way to embed this intuition, using more general priors (\autoref{tab:priors}) or task-specific priors  (\autoref{sec:priors}). For more complex problems where human intuition falls short, GCML can learn a prior. Then, the resulting human-interpretable graph of observations can inform the development of new, human-derived topological priors (\autoref{sec:mem_graphs}).

In our experiments, we use simple environments to demonstrate how model-dependent memory connectivity affects performance. Models like LSTM work nearly as well as GCM on problems like cartpole where a temporal prior makes sense (\autoref{fig:cartpole}), but the gap widens on the concentration environment where non-temporal priors are more suitable (\autoref{fig:memory}). The navigation ablation study (\autoref{fig:nav_edge}) demonstrates how using a suboptimal topological prior can negatively impact performance -- the dense prior (a fully-connected graph) performs nearly as poorly as the empty prior (no edges at all) in \autoref{fig:nav_edge}. Exploratory trials of combining human priors with learned priors resulted in performance greater than GCML alone, but not as good as GCM with human priors. Future work could evaluate these mixed priors on more complex problems over longer timescales, where human priors are less useful.

Across all experiments, GCM with human expertise received significantly more reward than the next best competitor. We believe that this is remarkable, considering that GCM uses notably fewer parameters than the other models (\autoref{fig:params}). GCML performance was generally less than LSTM, and most similar to the transformer performance across our experiments. Caveat emptor: we tackled simple tasks using smaller models, due to our limited computational capacity. These conclusions might not hold for those who train markedly larger models for billions of timesteps. Given more compute and longer episodes, we suspect the transformer and possibly GCML would outperform LSTM, like in \cite{Parisotto2019}.
\section{Related Work}

\paragraph{Memory in Reinforcement Learning}
We classify RNNs, MANNs, transformers, and related memory models as general memory. RNN-based architectures, such as long short-term memory (LSTM) \citep{hochreiter1997long} and the gated recurrent unit (GRU) \citep{chung2014empirical} are used heavily in RL to solve POMDP tasks \citep{Oh2016, Mnih2016, Mirowski2017}. RNNs update a recurrent state by combining an incoming observation with the previous recurrent state. Compared to transformers and similar methods, RNNs fail to retain information over longer episodes due to vanishing gradients \citep{Li2018}. By connecting relevant experiences directly and doing a single forward pass, GCM sidesteps the vanishing gradient issue.

MANNs address limited temporal range of RNNs \citep{Graves2014}. Unlike RNNs, MANNs have addressable external memory. The differentiable neural computer \citep{Graves2016} (DNC) is a fully-differentiable general-purpose computer that coined the term MANN. In the DNC, a RNN-based memory controller uses content-based addressing to read and write to specific memory addresses. The MERLIN MANN \citep{Wayne2018} outperformed DNCs on navigation tasks. The implementations of the MANNs are much more complex than RNNs. In contrast to transformers or RNNs, MANNs are much slower to train, and benefit from more compute. 

The transformer is the most ubiquitous implementation of self-attention \citep{Vaswani2017}. Until the gated transformer XL (GTrXL), transformers had mixed results in RL due to their brittle training requirements \citep{Mishra2018}. The GTrXL outperforms MERLIN, and by extension, DNCs in \cite{Parisotto2019}. When learning topological priors, our GCML borrows concepts from transformers such as positional encodings. The self-attention module in a transformer can be implemented using a single graph attention layer over a fully-connected graph \citep{joshi_2020}. Unlike self-attention in transformers, GCML is hard, sparse, and hierarchical. GCM does not use attention weights, nor a fully-connected graph.

Similar to our work, \cite{Savinov2018} build an observation graph, but specifically for navigation tasks, and do not use GNNs. \cite{Wu2019} use a probabilistic graphical model to represent spatial locations during indoor navigation. \cite{Eysenbach2019,Emmons2020} build a state-transition graph similar to our observation graph for model-based RL, but use A* and other methods to evaluate the graph.

\paragraph{Graph Neural Networks}
GNNs are most easily understood using a message-passing scheme \citep{Gilmer2017}, where each vertex in a graph sends and receives latent messages from its neighborhood. Each layer in the GNN learns to aggregate incoming messages into a hidden representation, which is then shared with the neighborhood. Convolutional graph neural networks \citep{Kipf2017} are a subcategory of GNNs and a generalization of convolutional neural networks (CNNs) to the graph domain. Convolutional GNNs tend to be efficient in both the computational and parameter sense due to their use of sliding filters and reliance on batched sums and matrix multiplies. 

Our method shares some similarities with graph attention networks \citep{Velickovic2018}, namely the pairwise-vertex MLP to compute edge significance. Graph RNNs \citep{Ruiz2020} are a generalization of RNNs to graph inputs with a fixed number of time-varying vertices, and tackle an entirely different problem than GCM. \cite{Chen2019,Li2019,Chen2020a} apply GNNs to RL for task-specific problems. \cite{Beck} implement feature aggregation for RL in a similar fashion to GNNs. The aggregated memory operator by \cite{Zweig2020} combines a GNN with a RNN to navigate a graph of states using reinforcement learning. To date, GCM is the only \emph{task-agnostic} RL memory model to utilize GNNs.
\section{Conclusion}
GCM provides a framework to embed task-specific priors into memory, without writing task-specific memory from the ground up. Embedding custom toplogical priors in the graph is trivial: it involves implementing a boolean function that determines whether or not observation $o_i$ is useful for decision making at time $t$. This simple feature makes GCM very versatile. Moreover, with GCML we showed that GCM can also be purposed to \textit{learn} topological priors \textit{without} human input, and in this case performs similarly to a transformer. When even basic domain knowledge is available (e.g., when the problem is spatial, or when it follows Newton's laws) GCM outperforms transformers, LSTMs, and DNCs, while using significantly fewer parameters. 

\section{Acknowledgements}
We gratefully acknowledge the support of Toshiba Europe Ltd., and the European Research Council (ERC) Project 949940 (gAIa). We thank Rudra Poudel, Jan Blumenkamp, and Qingbiao Li for their helpful discussions.

\clearpage
\bibliography{bib.bib}
\bibliographystyle{iclr2022_conference}

\clearpage
\appendix
\section{Interpreting Memory Graphs}
\label{sec:mem_graphs}
One of the strengths of GCM is that the memory is an interpretable graph. We can see precisely which past observations contribute to the current belief. In this section, we provide an example memory graph from the cartpole and navigation experiments using the learned topological prior (GCML). 

\begin{figure}[H]
    \centering
    \includegraphics[width=0.6\linewidth]{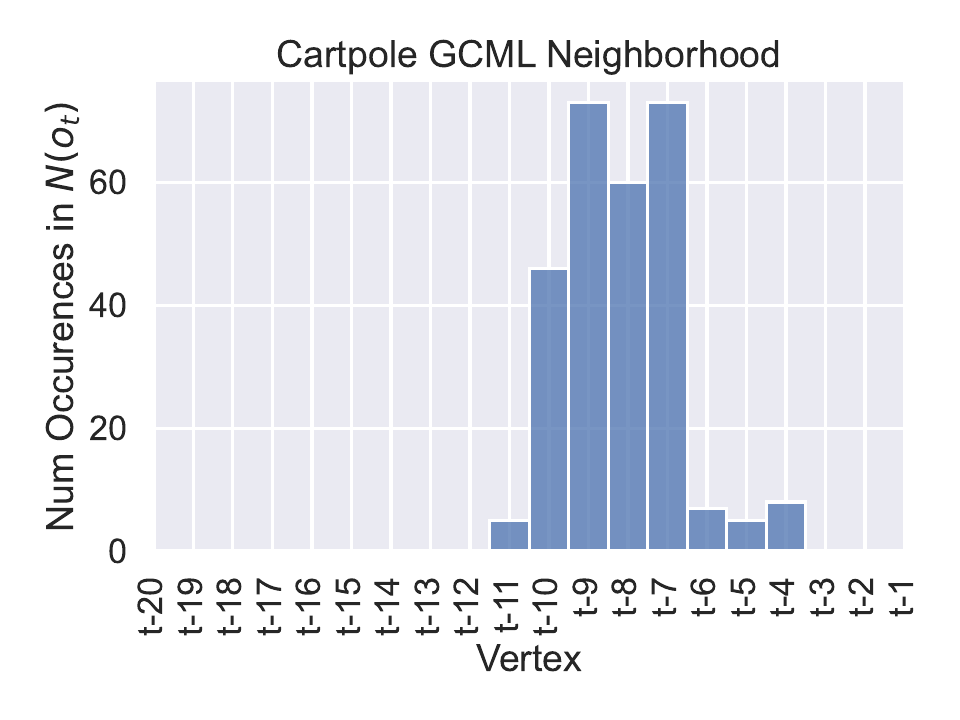}
    \caption{What do GCML neighborhoods look like for stateless cartpole? At each episodic timestep $t>20$, we record the neighborhood $N(o_t)$ GCML produces relative to $t$. We plot the accumulation of all neighborhoods over an episode, using $|z|=32, K=5$. We see that GCML learns a temporal prior, where each timestep uses observations from 7 to 10 timesteps ago. Surprisingly, GCML does not use the preceding observation, suggesting that the simulation contains high-frequency variations which prevent an accurate estimation of velocity between consecutive timesteps. Perhaps a human-defined prior with $\{t-7\}$ would produce a smoothed signal, leading to better performance than our $\{t-1, t-2\}$ prior. With $K=5$, the maximum possible neighborhood size is 5. The mean neighborhood size is $1.54$, demonstrating that GCML can learn a peaked distribution, resulting in sparse graphs.}
    \label{fig:cartpole_edges}
\end{figure}

\begin{figure}[H]
    \centering
    \includegraphics[width=0.6\linewidth]{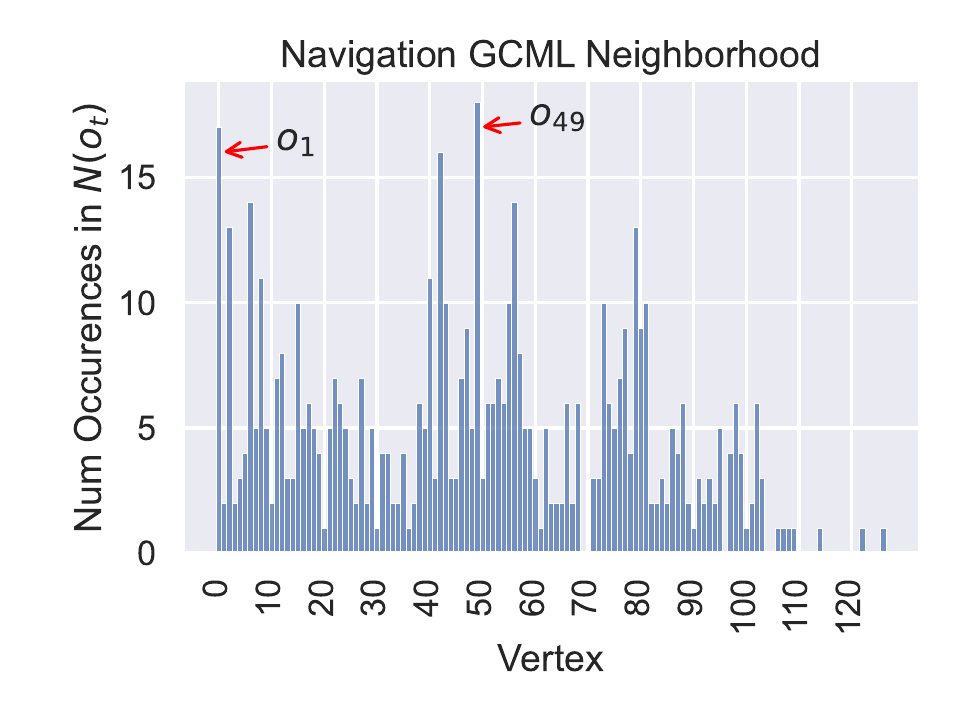}
    \caption{What do GCML navigation neighborhoods look like? At each episodic timestep, we record the neighborhood $N(o_t)$ GCML produces. We plot the accumulation of all neighborhoods over an episode, using $|z|=32, K=5$. Unlike the \autoref{fig:cartpole_edges}, vertices here are labeled in an absolute fashion (when they occurred). We see there are a few ``key'' observations (red arrows), such as the start vertex $o_1$ or $o_{49}$, similar to the use of keyframes in visual SLAM. Unlike stateless cartpole, information in the navigation task is spread over many observations. With $K=5$, the maximum neighborhood size is 5 and the mean neighborhood size is $4.23$, illustrating a flat distribution, and resulting in a denser graph.}
    \label{fig:nav_hist}
\end{figure}

\begin{figure}[H]
    \centering
    \begin{subfigure}{0.6\linewidth}
        \includegraphics[width=\linewidth]{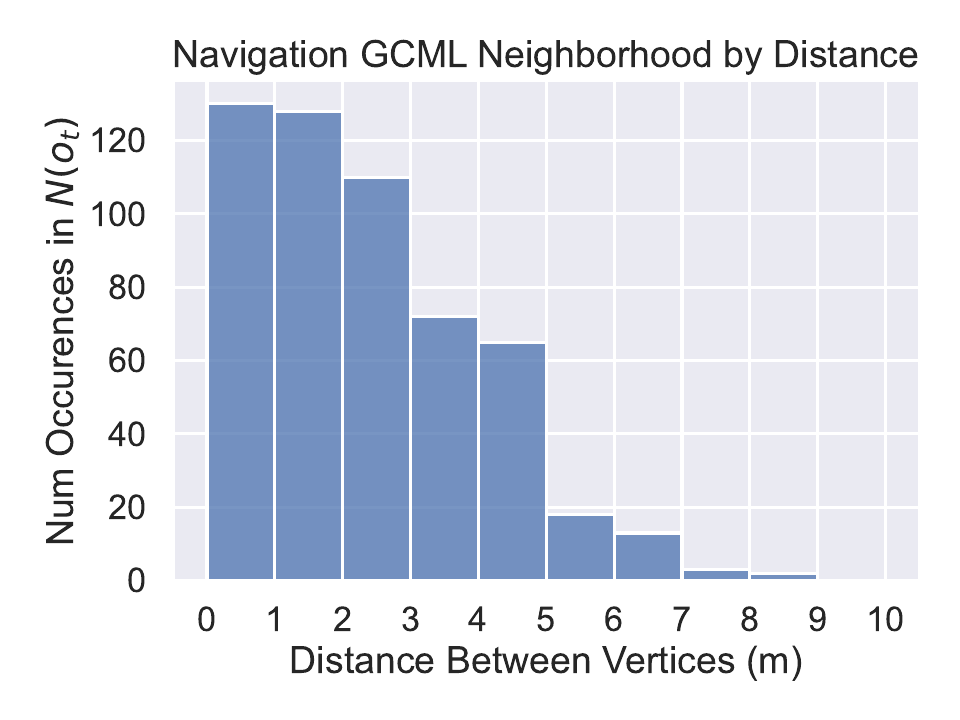}
        \caption{}
        \label{fig:nav_distance_left}
    \end{subfigure}
    \begin{subfigure}{0.6\linewidth}
        \includegraphics[width=\linewidth]{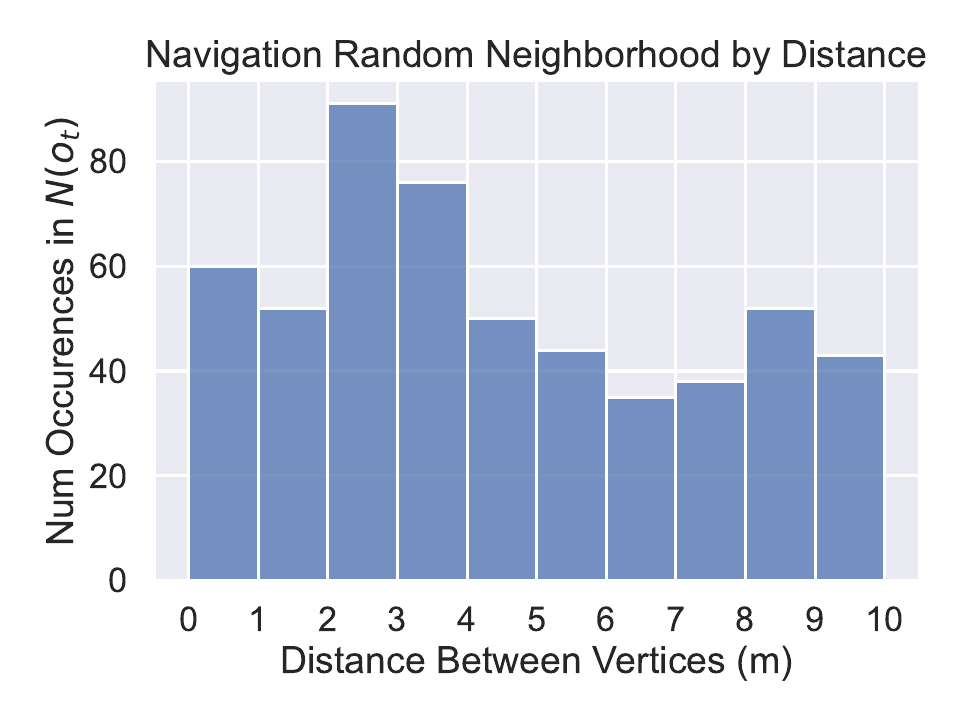}
        \caption{}
        \label{fig:nav_distance_right}
    \end{subfigure}
    \caption{What do GCML navigation neighborhoods look like in the spatial domain? (\subref{fig:nav_distance_left}) At each episodic timestep, we record the neighborhood $N(o_t)$ GCML produces. We plot the accumulation of all neighborhoods over an episode, using $|z|=32, K=5$. We bin each vertex in the neighborhood $N(o_t)$ by its distance to the vertex $o_t$. We see GCML looks to be spatially-biased, heavily prioritizing nearby vertices over further vertices. (\subref{fig:nav_distance_right}) We shuffle the edge indices from \subref{fig:nav_distance_left} to see if the task itself is biased towards shorter distances (e.g. maybe agent spends lots of time in a single room 4m\textsuperscript{2} in size, and all vertices are within 4m). We find this is not the case, and that the vertices span a large distance. \subref{fig:nav_distance_left} and \subref{fig:nav_distance_right} together prove that GCML indeed learns a spatial prior.}
    \label{fig:nav_distance}
\end{figure}

\begin{figure}[H]
    \centering
    \begin{subfigure}{\linewidth}
    \includegraphics[width=\linewidth]{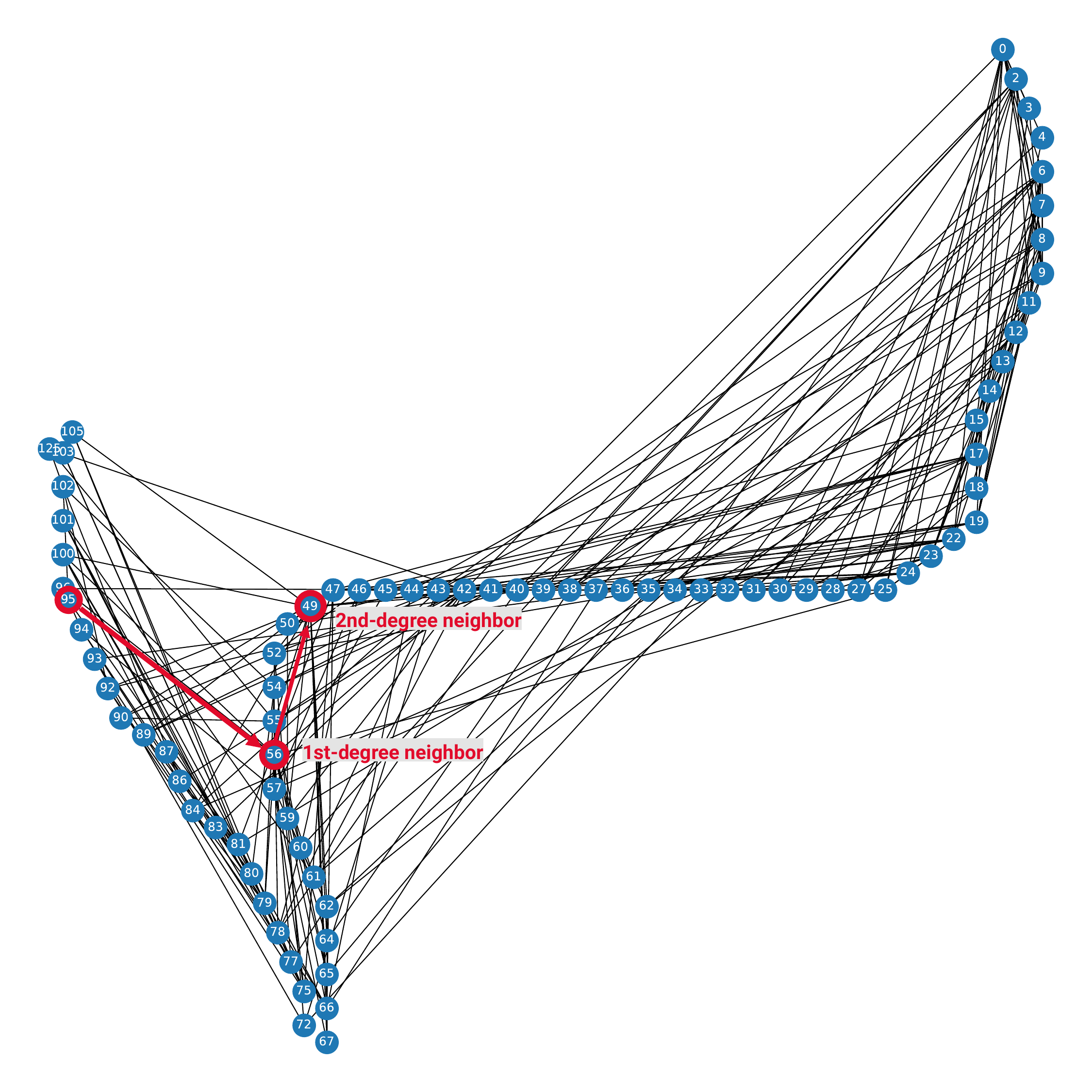}
        \subcaption{}
        \label{fig:nav_graph}
    \end{subfigure}
    
    \begin{subfigure}{0.3\linewidth}
    \includegraphics[width=\linewidth]{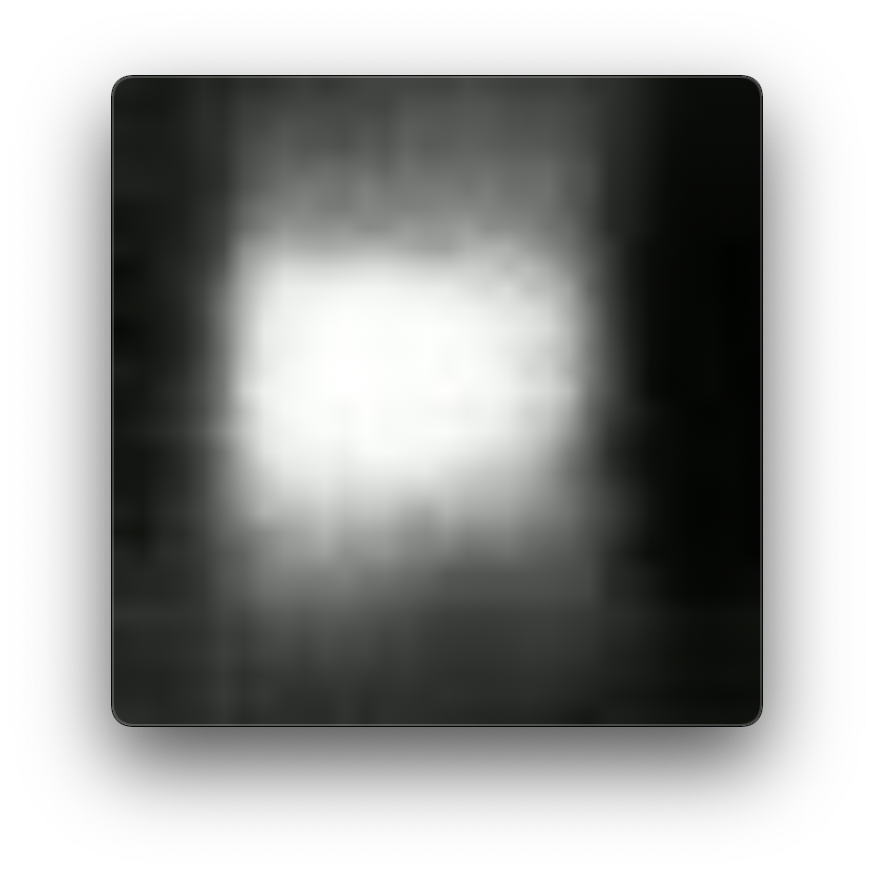}
        \subcaption{}
        \label{fig:nav_graph_left}
    \end{subfigure}
    \begin{subfigure}{0.3\linewidth}
    \includegraphics[width=\linewidth]{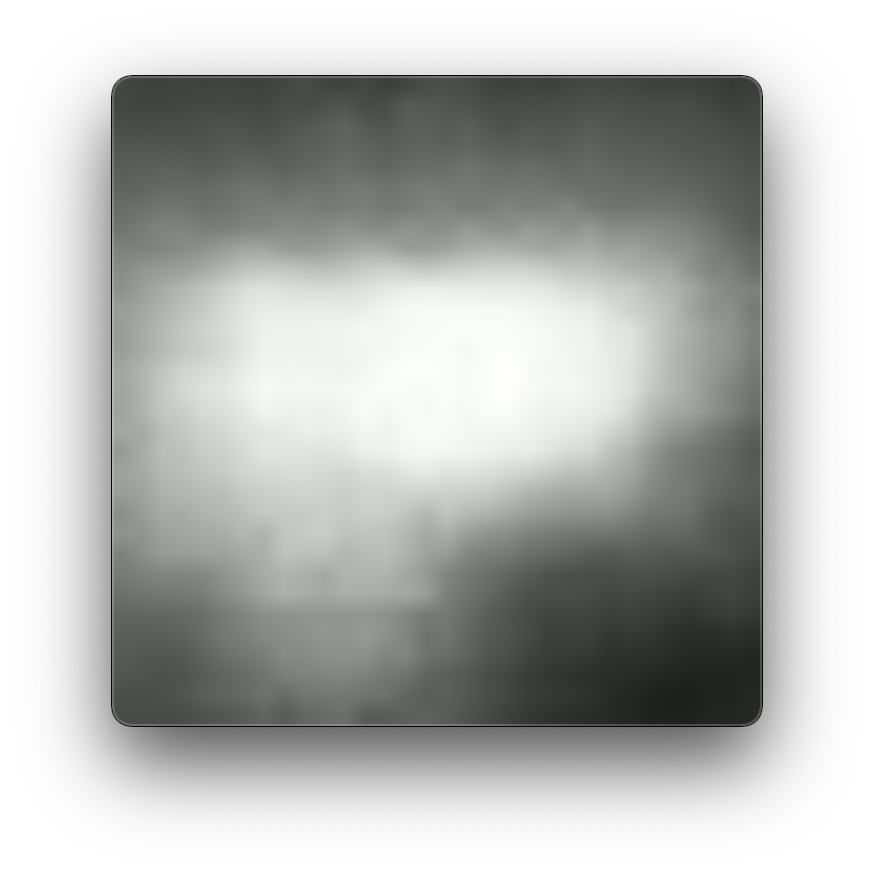}
        \subcaption{}
        \label{fig:nav_graph_right}
    \end{subfigure}
    \caption{(\subref{fig:nav_graph})~A visual representation of the memory graph from the episode used in \autoref{fig:nav_hist} and \autoref{fig:nav_distance}. Vertices are labeled by timestep and projected into a 2D-plane based on their physical location. Overlapping vertices (e.g. when the agent rotates in place) are not shown. Here, we find clues for why $o_{49}$ was so heavily favored in \autoref{fig:nav_hist}. We reconstruct the $o_{49}$ depth image and find it was precisely when the agent left a narrow corridor (\subref{fig:nav_graph_left}) to enter a large living room (\subref{fig:nav_graph_right}). As it explores the living room ($o_{50}-o_{94}$), it localizes itself relative $o_{49}$, using past information to exit through a different door than it entered ($o_{95}$), maximizing exploration reward. We trace the belief at $o_{95}$ through first-degree neighbor $o_{56}$ to second-degree neighbor $o_{49}$ in red. This introspection leads us to believe we should pay attention to doorways when designing topological priors for navigation.}
\end{figure}

\clearpage
\section{Topological Prior Examples}
\label{sec:eg_priors}
To demonstrate how simple it is to write topological priors, we provide two examples of task specific priors in Pytorch. First, imagine a hospital agent where key observations occur when the medicines administered to the patient change. This could be implemented as such
\begin{lstlisting}[language=Python]
import torch
class MedicalPrior(torch.nn.Module):
  def forward(self, V):
    # V is V_t (contains o_t)
    meds = get_meds_from_observation(V[:-1])
    N = torch.where(meds != meds.roll(1,0))
    return N
\end{lstlisting}
where \texttt{get\_meds\_from\_observation} slices a tensor of observations to extract the medications administered at each timestep. Next, assume we are learning a dynamics model for an aircraft in flight. We want to exclude all past observations after a bird strike takes out an engine, and learn a new reduced-capability dynamics model
\begin{lstlisting}[language=Python]
import torch
class AircraftPrior(torch.nn.Module):
  max_deviation = 8 # meters per second squared
  def forward(self, V):
    errors = get_dyn_model_error(V[:-1]) # Obs err using dyn model
    latest_damage_event = (errors > max_deviation).nonzero()[-1,0]
    N = torch.range(latest_damage_event, V.shape[0])
    return N
\end{lstlisting}
where \texttt{get\_dyn\_model\_error} returns the acceleration disparity between a learned dynamics model and sensor measurements for each timestep.

\clearpage
\section{Environmental Details}
\label{sec:exp_details}
\subsection{Stateless Cartpole}
We do not change any of the default environment settings from the OpenAI defaults: we use an episode length of $200$, with a reward of $1$ for each timestep survived. The episode ends when the pole angle is greater than 12 degrees or the cart position is further than 2.4m from the origin. Actions correspond to a constant force in either the left or right directions. We use the OpenAI gym definition (mean reward of 195) \cite{Brockman2016} to determine if the agent is successful. The value loss coefficient hyperparameter was used from the RLlib cartpole example, and thus differs from the non-cartpole default.

\subsection{Concentration Game}
The agent is given $n/2$ pairs of shuffled face-down cards, and must flip two cards face up. If the cards match, they remain face up, otherwise they are turned back over again. Once the player has matched all the cards, the game ends. We model the game of memory using a \emph{pointer}, which the player moves to read and flip cards (\autoref{fig:sub_memory}). The observation space consists of the pointer (card index and card value) and the last flipped (if any) face-up card. Cards are represented as one-hot vectors. The agent receives a reward for each pair it matches, with a cumulative reward of one for matching all the cards.

\subsection{Navigation}
The navigation experiment operated on the CVPR Habitat 2020 challenge \citep{Savva2019} validation scene from the MP3D dataset \citep{Chang2018}. We used the same list of agent start coordinates as used during the challenge. $32 \times 32$ depth images with range $[0.5, 5\textrm{m}]$ and a 79 degree field of view were compressed into a 64-dimensional latent representations using a 6 layer (3 encoder, 3 decoder) convolutional $\beta$-VAE \cite{Kingma2014} with $\beta=0.01$, ReLU activation, and batch normalization. The $\beta$-VAE was pretrained until convergence using random actions. The full observations consists of agent heading and coordinates relative to the current start location, the previous action, and the latent VAE representation. The agent can rotate 30 degrees in either direction or move 0.25m forward. The agent receives a reward of 0.01 for exploring a new area of radius 0.20m. 

\clearpage
\section{Experiment Hyperparameters}
\label{sec:hyperparams}
The following hyperparameters used for each experiment across all memory modules. We generally reduce the learning rate as well as increase the batch size to produce more consistent reward curves. Nearly all other hyperparameters are Ray RLlib defaults.
\begin{center}
        \begin{tabular}{ccc}
            \toprule
            Term & Value & RLlib Default \\
            \toprule
            \multicolumn{3}{c}{Navigation IMPALA}\\
            \midrule
            Decay factor $\gamma$ & 0.99 & \checkmark\\
            Value function loss coef. & 0.5 & \checkmark\\
            Entropy loss coef. & 0.001 & 0.01 \\
            Gradient clipping & 40 & \checkmark\\
            Learning rate & 0.005 & \checkmark\\
            Num. SGD iters & 1 & \checkmark\\
            Experience replay ratio & 1:1 & 0:1\\
            Batch size & 1024 & 500\\
            GAE $\lambda$ & 1.0 & \checkmark\\
            V-trace $\rho$ & 1.0 & \checkmark\\
            \midrule
            \multicolumn{3}{c}{Cartpole PPO}\\
            \midrule
            Decay factor $\gamma$ & 0.99 & \checkmark\\
            Value function loss coef. & 1e-5 & \checkmark (cartpole-specific)\\
            Entropy loss coef. & 0.0 & \checkmark\\
            Value function clipping & 10.0 & \checkmark\\
            KL target & 0.01 & \checkmark\\
            KL coefficient & 0.2 & \checkmark\\
            PPO clipping & 0.3 & \checkmark\\
            Value clipping & 0.3 & \checkmark\\
            Learning rate & 5e-5 & \checkmark\\
            Num. SGD iters & 30 & \checkmark\\
            Batch size & 4000 & \checkmark\\
            Minibatch size & 128 & \checkmark\\
            GAE $\lambda$ & 1.0 & \checkmark \\
            \midrule
            \multicolumn{3}{c}{Concentration PPO}\\
            \midrule
            Decay factor $\gamma$ & 0.99 & \checkmark\\
            Value function loss coef. & 1.0 & \checkmark\\
            Entropy loss coef. & 0.0 & \checkmark\\
            Value function clipping & 10.0 & \checkmark\\
            KL target & 0.01 & \checkmark\\
            KL coefficient & 0.2 & \checkmark\\
            PPO clipping & 0.3 & \checkmark\\
            Value clipping & 0.3 & \checkmark\\
            Learning rate & 3e-4 & 5e-5\\
            Num. SGD iters & 30 & \checkmark\\
            Batch size & 4000 & \checkmark\\
            Minibatch size & 4000 & 128\\
            GAE $\lambda$ & 1.0 & \checkmark \\
            \bottomrule
        \end{tabular}
        \label{tab:hyperparams}
\end{center}

\clearpage

\end{document}